\documentclass{article} % For LaTeX2e
\usepackage{iclr2026_conference,times}
\usepackage{graphicx}

\usepackage{amsmath,amsfonts,bm}

\def\eqref#1{equation~\ref{#1}}
\def\1{\bm{1}}

\DeclareMathAlphabet{\mathsfit}{\encodingdefault}{\sfdefault}{m}{sl}
\SetMathAlphabet{\mathsfit}{bold}{\encodingdefault}{\sfdefault}{bx}{n}

\usepackage{hyperref}
\usepackage{url}

\title{A Cortically Inspired Architecture for Modular Perceptual AI}

\author{Prerna Luthra \\
Independent Researcher\\
\texttt{prerna@samvedna.ai} \\
}

\iclrfinalcopy % Uncomment for camera-ready version, but NOT for submission.

\begin{document}

\maketitle

\begin{abstract}
This paper bridges neuroscience and artificial intelligence to propose a cortically inspired blueprint for modular perceptual AI. While current monolithic models such as GPT-4V achieve impressive performance, they often struggle to explicitly support interpretability, compositional generalization, and adaptive robustness - hallmarks of human cognition. Drawing on neuroscientific models of cortical modularity, predictive processing, and cross-modal integration, we advocate decomposing perception into specialized, interacting modules. This architecture supports structured, human-inspired reasoning by making internal inference processes explicit through hierarchical predictive feedback loops and shared latent spaces. Our proof-of-concept study provides empirical evidence that modular decomposition yields more stable and inspectable representations. By grounding AI design in biologically validated principles, we move toward systems that not only perform well, but also support more transparent and human-aligned inference.

\end{abstract}

\section{Introduction}
Modern perceptual AI systems spanning computer vision, speech recognition, and multimodal understanding are overwhelmingly built as large monolithic models trained end-to-end \citep{Krizhevsky2012AlexNet, Devlin2019BERT}. This monolithic paradigm has delivered remarkable achievements, but it comes with well-documented limitations. Deep networks often operate as opaque ``black boxes" that require enormous training data, and can struggle to generalize beyond the narrow statistical patterns of their training distribution \citep{Marcus2018Critique, Lake2017HumanLike}. In particular, such models show brittleness in out-of-distribution scenarios and offer limited transparency or modularity of internal inference. As argued in a public exchange between Gary Marcus and Yoshua Bengio, ``expecting a monolithic architecture to handle abstraction and reasoning is unrealistic". However, much of today’s AI, including the latest foundation models \citep{Bommasani2021FoundationModels}, continues to pursue ever larger end-to-end networks as a one-size-fits-all solution. 

These limitations motivate the use of explicit cognitive organization as an architectural prior. We posit that the next generation of AI should move beyond unified black-box networks toward modular, cortically inspired systems. In biological brains, perception emerges from an interplay of specialized cortical regions (visual, auditory, language areas, etc.) organized into a deep hierarchy of processing stages \citep{Felleman1991Hierarchy}. Information flows not just bottom-up but also top-down, with higher cortical areas continuously generating predictions to inform lower-level processing—a principle known as predictive coding \citep{Rao1999PredictiveCoding, Friston2005CorticalResponses}. This predictive and hierarchical organization enables efficient data usage, generalization, and integration of multiple modalities. Indeed, the brain is often described as a ``prediction machine" that actively anticipates sensory inputs \citep{Clark2013PredictiveBrains}. These insights from neuroscience motivate us to rethink AI architectures. Instead of colossal homogeneous networks, we advocate modular designs composed of interacting components that mirror the brain’s division of labor and recurrent predictive loops.

Some early steps toward modular perceptual AI have shown promise. For example, neural module networks for vision–language tasks demonstrate how separate learned sub-networks can be composed to solve complex queries \citep{Andreas2016NMN}. Similarly, mixture-of-experts models and recurrent independent mechanisms introduce sparsely activated sub-networks aimed at capturing the benefits of specialization \citep{Goyal2019RIMs}. However, these approaches often remain trained end-to-end and lack a unifying cognitively grounded framework that explains how specialization, coordination, and feedback should be structured. In contrast, cortical biology suggests a more principled architectural blueprint: a network of predictive, sparsely communicating components, each specialized for a particular modality or feature, yet operating under a shared predictive coding paradigm \citep{Rao1999PredictiveCoding, Friston2005CorticalResponses}. In human perception, multimodal processing arises from such an organization - for example, visual and auditory cortices process inputs separately but converge in higher association areas that resolve cross-modal predictions, allowing flexible integration without collapsing all signals into a single undifferentiated representation \citep{Beauchamp2004STS, Calvert2000Crossmodal}

Building on these neuroscientific observations, we propose three core principles for designing cortically inspired perceptual AI systems:

\paragraph{Modular Specialization and Hierarchical Organization.} Perceptual AI should be factorized into semi-independent modules specialized for distinct tasks or modalities (e.g., vision, language, audio subsystems), analogous to functional cortical areas and organized hierarchically to support abstraction and contextual feedback \citep{Mountcastle1997Columns, Kanwisher2010Specificity, Felleman1991Hierarchy}.

\paragraph{Predictive Processing.} Rather than purely feed-forward computation, systems should leverage recurrent predictive mechanisms in which components generate top-down expectations and update representations via prediction errors\citep{Rao1999PredictiveCoding, Friston2005CorticalResponses}, supporting robustness and uncertainty-sensitive inference \citep{Clark2013PredictiveBrains}.

\paragraph{Cross-Modal Integration.} While modular, perceptual AI should also be integrative. In the brain, distinct cortical regions interact through structured connectivity to produce coherent percepts \citep{Sporns2016ModularBrain}. Analogously, specialized AI modules should communicate through well-defined interfaces, allowing distinct modalities to contribute complementary information to shared inference processes. Shared embedding spaces \citep{Radford2021CLIP, Girdhar2023ImageBind} or association-like representations can help align representations across modalities, enabling flexible multimodal reasoning while preserving modular structure. 

In this paper, we advance the case for cortically inspired modular perception and make four contributions. (1) We synthesize neuroscientific and cognitive evidence motivating modular specialization, predictive feedback, and cross-modal integration. (2) We outline an architectural blueprint that operationalizes these principles in a modular perceptual system. (3) We present a focused diagnostic proof-of-concept study showing that explicit semantic modularization of internal representations within a large language model improves within-domain feature stability without sacrificing reconstruction fidelity. (4) Finally, we situate this proposal relative to monolithic, mixture-of-experts, and neuro-symbolic approaches, highlighting its implications for interpretability and robustness.

\section{Neuroscientific Motivation}
To overcome the limitations of monolithic AI systems, such as brittleness, opacity, and inflexibility, we draw on foundational neuroscientific principles of cortical function. This section synthesizes empirical and conceptual evidence for three core design tenets: modular specialization, predictive feedback, and cross-modal integration. Each principle maps directly onto cortical organization and offers actionable guidance for developing interpretable, and adaptive AI systems.

\subsection{Modular Specialization and Isolation of Function}
The mammalian cortex is composed of semi-autonomous modules. Seminal work in neuroscience demonstrates that different cortical regions specialize in distinct functional domains. For instance, the fusiform face area is dedicated to face perception, MT/V5 to motion processing, and V4 to color processing - each revealing finely tuned cortical maps \citep{Kanwisher2010Specificity, Zeki1993Vision}. The concept of the cortical column as a canonical microcircuit was famously introduced by \citet{Mountcastle1997Columns}, emphasizing the brain’s modular construction.
Higher-order cortical association areas also show lateralized functional specialization - for instance, the left prefrontal cortex for language and the right for visuospatial tasks. Functional connectivity studies further show that the cortex exhibits hierarchically modular organization: subnetworks for sensory, motor, and cognitive functions operate largely independently but interact through structured interfaces \citep{Sporns2016ModularBrain}.
This kind of architectural separation in the brain is associated with robustness, interpretability, and continual learning. Modular organization allows for task-specific updates without global interference, mitigating catastrophic forgetting - a chronic problem in monolithic models. For example, \citet{Ellefsen2015Modularity} demonstrated that modular artificial neural networks can outperform monolithic ones in learning new tasks while preserving previously acquired knowledge. In contrast, widely used monolithic architectures such as large transformers tend to represent multiple functions within a shared parameter space, leading to entangled internal representations. This limits transparency and makes targeted optimization difficult, as localized updates can produce unintended downstream effects. AI systems inspired by neural modularity, such as Neural Module Networks \citep{Andreas2016NMN}, demonstrate that explicitly separating functional components can support greater interpretability and task efficiency.

\subsection{Predictive Feedback and Active Inference}
A defining feature of cortical computation is pervasive recurrent feedback. Predictive coding theory posits that the brain continuously generates top-down predictions and updates them through bottom-up prediction errors \citep{Rao1999PredictiveCoding, Friston2005CorticalResponses}. Feedback loops refine perception iteratively, producing a dynamic equilibrium between expectation and sensation.
Experimental evidence supports this framework. For instance, visual cortex activity (e.g., V1) is suppressed for expected stimuli \citep{Alink2010Predictability}, and auditory cortex shows anticipatory activation in response to cued stimuli \citep{Sohoglu2012SpeechPrediction}. \citet{Clark2013PredictiveBrains} famously described the brain as a ``prediction machine" that reduces surprise via recursive inference. These circuits enable context-sensitive, resilient interpretation of noisy or incomplete inputs.
Most AI models today are feed-forward and lack the ability to refine hypotheses post hoc. This architectural limitation is thought to contribute to hallucinations: confident but incorrect outputs. Unlike the brain, which checks its internal generative outputs against sensory input, AI models lack embedded feedback paths.
Incorporating cortical feedback principles into AI could reduce these hallucinations. Feedback-enhanced architectures like recurrent vision transformers \citep{Pan2022ViTDet} and recurrent independent mechanisms \citep{Goyal2019RIMs} show promise in handling ambiguity through iterative refinement. These observations motivate a more detailed treatment of hallucinations as emergent artifacts of predictive inference, which we revisit in the architectural section.

\subsection{Cross-Modal Integration and Semantic Convergence}
The brain integrates multimodal data into unified percepts. Association regions such as the superior temporal sulcus (STS) and posterior parietal cortex (PPC) bind information from vision, audition, and language into coherent representations \citep{Beauchamp2004STS, Calvert2000Crossmodal}. This fusion enhances perception, especially under uncertainty, and operates through temporal coincidence and semantic congruence.
Examples include the McGurk effect, where visual cues alter auditory perception, and predictive cross-activation - e.g., speech primes activity in visual cortex \citep{Summerfield2009Expectation}. Such deep integration is present even in early sensory areas, suggesting an early and recursive multimodal architecture.
Most current AI systems treat modalities in isolation or merge them through primarily static embedding alignment (e.g., CLIP-style contrastive objectives \citep{Radford2021CLIP}). While recent systems such as ImageBind extend this alignment across many modalities \citep{Girdhar2023ImageBind}, they lack explicit reciprocal predictive links across modalities. To more fully emulate brain-like perception, architectures should include modality-specific modules linked through a predictive workspace, allowing cross-modal error checking and redundancy, which are vital for grounding and interpretability.
In systems without dynamic, inference-time grounding, hallucinations often stem from language priors overpowering perceptual evidence. Recent studies show that strengthening the influence of visual input during generation in vision–language models reduces hallucination rates \citep{Favero2024Hallucination}, reinforcing the cortical principle that multimodal grounding serves as a safeguard against spurious outputs.

\section{Architectural Blueprint for Modular Cortical AI}
Inspired by neuroscience, we propose an AI architecture that decomposes perception and reasoning into specialized, interacting modules. This design operationalizes three cortical principles – modular specialization, cross-modal integration, and predictive feedback – to enable more robust and interpretable intelligence than monolithic networks. The key components of this architectural blueprint, along with their functional roles and interactions, are described below. Figure~1 provides a schematic illustration of the proposed system. 
\begin{figure}[h]
    \centering
    \includegraphics[width=\linewidth]
    {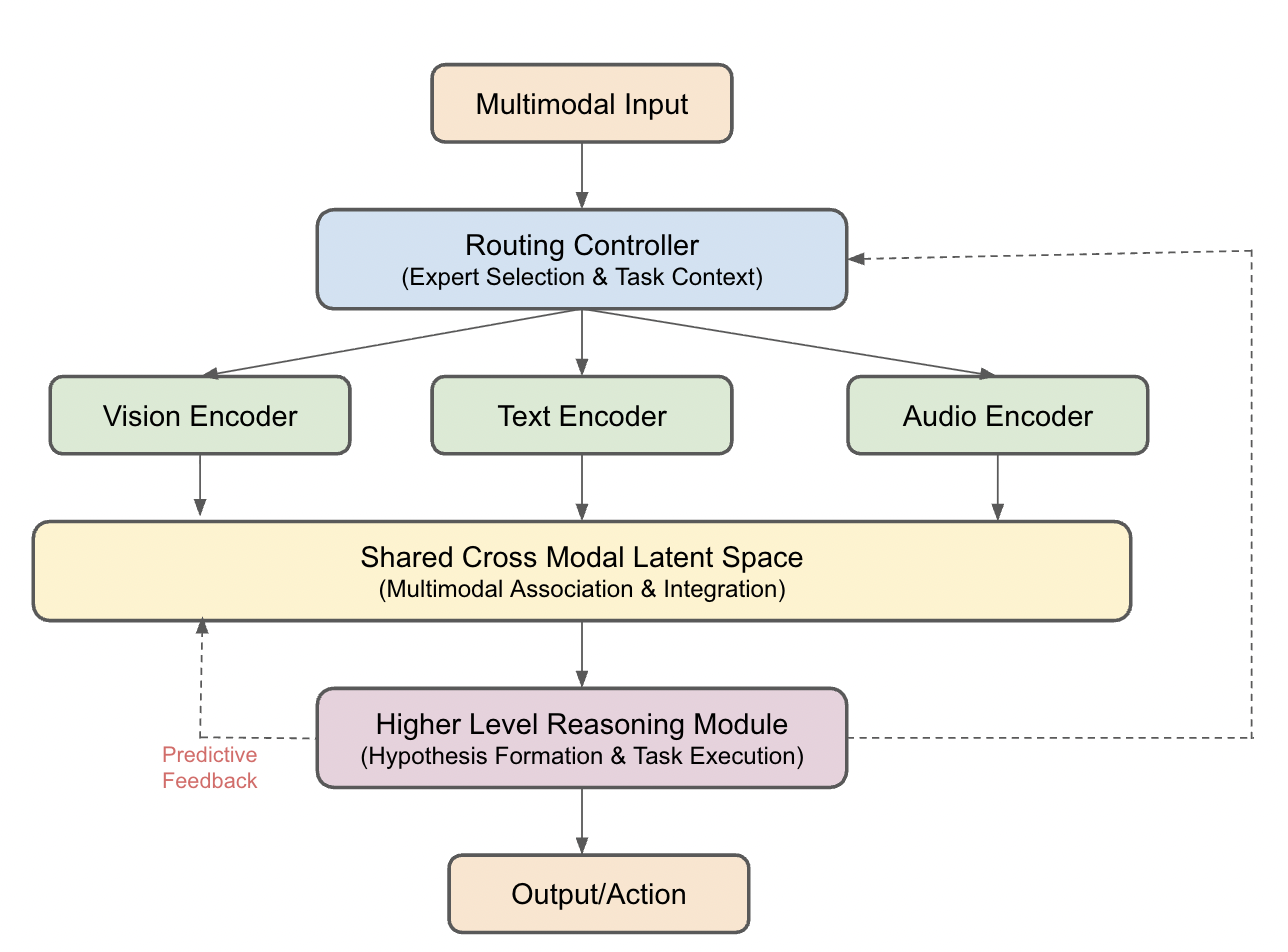}
    \caption{Architectural blueprint for modular perceptual AI, illustrating specialist encoders, a shared multimodal workspace, routing control, and predictive feedback loops.}
    \label{fig:cortical-ai}
\end{figure}

\subsection{Specialist Encoders for Each Modality}
At the system’s periphery, we employ specialist encoder modules dedicated to particular input modalities or tasks, forming the perceptual front-end of the architecture. Each encoder can be instantiated as a pre-trained expert network that transforms raw sensory data into a latent representation optimized for its domain. For example, Whisper can serve as a speech-audio expert for mapping acoustic input into linguistic representations \citep{Radford2022Whisper}. Similarly, vision encoders (e.g., convolutional or ViT-based models) transform images into structured visual representations, while large language models such as LLaMA or Vicuna operate as text-based reasoning experts. This separation mirrors cortical specialization, where early sensory areas compute modality-specific representations prior to downstream integration.
This modular approach mirrors the brain’s division into areas specialized for vision, audition, language, etc. Each module can be trained independently, improving robustness and interpretability: a failure in one module does not destabilize the whole, and each expert’s output can be inspected and debugged independently. New capabilities can be added by plugging in a new module without retraining the entire network.

\subsection{Shared Cross-Modal Latent Space}
To enable coordination across specialized modules, encoder outputs are mapped into a shared latent space inspired by multimodal cortical association areas. This space supports semantic alignment across modalities while preserving modular processing at the periphery.
Existing systems provide useful precedents for such shared representations. For example, CLIP jointly trains vision and text encoders to align embeddings of image–caption pairs within a common semantic space  \citep{Radford2021CLIP}. ImageBind extends this idea to multiple modalities (e.g., depth and audio) by aligning them through shared visual supervision \citep{Girdhar2023ImageBind}.
In our framework, the shared latent space functions as a dynamic workspace for cross-modal semantic convergence rather than a static alignment layer. It enables zero-shot transfer across modalities, supports direct cross-modal comparisons (e.g., matching sounds to images), and provides a locus for integrating top-down predictions and bottom-up evidence. This design mirrors the role of cortical association regions, which coordinate information across sensory systems while enabling flexible, context-dependent integration.

\subsection{Routing Controller for Expert Selection}
A routing controller governs which specialist modules are engaged for a given input, task, or intermediate inference state. This component is inspired by the brain’s ability to flexibly allocate processing across cortical regions based on context and behavioral goals, rather than relying on a fixed processing pipeline.
From a computational perspective, this controller may draw on mechanisms developed in Mixture-of-Experts (MoE) models, where a sparse subset of experts is activated per input \citep{Shazeer2017MoE, Lepikhin2020GShard}, or on efficient variants such as the Switch Transformer, which selects a single expert per token \citep{Fedus2021Switch}. However, unlike standard MoE routing, the controller here operates at the level of modular reasoning and perception rather than purely for computational efficiency.
Routing decisions can be informed by input modality, task context, or learned properties of the shared latent representation. For example, an image input may first activate a vision encoder, whose latent representation is then passed to a language reasoning module. The language module may, in turn, issue a query that recruits an audio expert or other specialized component. By making such control decisions explicit, this modular orchestration exposes the reasoning chain for inspection while enabling scalability through selective activation rather than monolithic parameter growth.

\subsection{Recurrent Predictive Feedback Loops}
The final architectural ingredient is recurrent feedback across modules. Drawing on predictive coding theories, the system incorporates top-down loops in which higher-level representations generate predictions that constrain lower-level processing \citep{Lee2003BayesianVision, Friston2005CorticalResponses}.
For example, a language-based inference module may form expectations about semantic or contextual properties of an input (e.g., the likely language being spoken), which can bias downstream auditory processing toward compatible interpretations. Through such iterative refinement, modules can resolve ambiguity and improve robustness under noisy or incomplete sensory conditions. Recent architectures such as Recurrent Independent Mechanisms (RIMs) \citep{Goyal2019RIMs} and Recurrent Vision Transformers \citep{Pan2022ViTDet} demonstrate how sparse, recurrent interactions allow modules to update their representations based on context and feedback.
From this perspective, hallucinations can be reframed as provisional generative hypotheses rather than one-shot failures. Instead of emitting a single unconstrained output, a feedback-driven system can iteratively propose, evaluate, and revise predictions until they achieve consistency across modalities and levels of abstraction. This process resembles hypothesis testing in biological perception, where interpretations are continuously adjusted in light of new evidence.

\subsection{Information Flow and Feedback-Driven Reasoning}
The system operates through an iterative loop: encode → coordinate → hypothesize → feedback. Sensory data are first processed by specialist encoder modules and mapped into a shared latent workspace, where representations from different modalities can be coordinated without collapsing modular structure. A higher-level reasoning module interprets this joint state to form provisional hypotheses about the input and task context.
These hypotheses are then broadcast as top-down contextual signals, prompting lower-level modules to re-evaluate their representations in light of global expectations. Each cycle propagates prediction errors across levels, enabling iterative refinement and increased cross-modal consistency. The routing controller facilitates this process by activating relevant modules and supporting repeated interaction until representations stabilize.
Through this feedback-driven organization, the architecture supports interpretable and adaptive inference while remaining structurally modular and cognitively grounded, reflecting core principles of cortical information processing.

\subsubsection{Hallucinations as Predictive Artifacts of Generative Inference}
From a cortical perspective, hallucinations can be understood as byproducts of generative inference rather than purely as errors. Predictive coding and the free-energy principle posit that perception operates by minimizing surprise through continuous prediction error reduction \citep{Friston2010FreeEnergy}. When top-down predictions dominate weak or ambiguous sensory input, internally generated hypotheses may persist, manifesting as hallucinations. Closely related phenomena occur in biological systems, including dreaming, expectation-driven illusions, and perceptual filling-in \citep{Lee2003BayesianVision, Hobson2014Dreams}, and clinical hallucinations associated with strong priors and weak sensory evidence \citep{Sterzer2018PredictiveCodingPsychosis}.

Under this view, hallucinations in AI emerge as a predictable consequence of adaptive inference under uncertainty rather than as pathological failures. In modular, feedback-driven systems, such generative hypotheses need not be emitted uncritically. Instead, they can be iteratively evaluated and constrained through recurrent feedback, cross-modal consistency checks, or external grounding signals. Analogous to imagination in biological cognition, internally generated predictions can support creativity and hypothesis formation, while verification mechanisms ensure alignment with sensory evidence and task constraints. Because hypotheses and feedback pathways are modular and explicit, the sources and resolution of hallucinations become inspectable rather than opaque.

\section{Relation to Existing Architectures}
\subsection{High-Performing Monolithic Models and End-to-End Successes}
Proponents of monolithic AI systems point to the remarkable performance of large end-to-end models across a wide range of tasks. OpenAI’s GPT-4, for example, is a single transformer-based architecture that demonstrates strong reasoning and language capabilities on numerous benchmarks, including professional and academic examinations \citep{OpenAI2023GPT4}. Similarly, Google DeepMind’s Gemini, a multimodal successor to PaLM-2, achieves competitive results across reasoning, coding, and long-context understanding tasks. In the vision–language domain, models such as Flamingo show that unified architectures can achieve strong few-shot performance on multimodal benchmarks without explicit modular decomposition \citep{Alayrac2022Flamingo}. Collectively, these systems highlight the empirical effectiveness of large, general-purpose models trained end-to-end.

At the same time, despite their performance, such models are not explicitly designed around established cognitive or neuroscientific principles. Their internal computations remain highly entangled and opaque, motivating interest in complementary architectural paradigms that prioritize modularity, interpretability, and principled inference grounded in models of human cognition.

\subsection{Hybrid Architectures and Neuro-Symbolic Frameworks}
Between fully monolithic and fully modular approaches, a range of hybrid architectures has emerged that blends specialization with end-to-end learning. Neural Module Networks (NMNs), for example, explicitly compose neural modules corresponding to sub-tasks such as filtering, counting, or relational reasoning. Early NMN systems for visual question answering demonstrated that parsing a query into a layout of sub-tasks enables the dynamic assembly of task-specific networks from a shared module library, yielding improved interpretability and compositional generalization \citep{Andreas2016NMN}. However, these systems typically depend on external mechanisms, such as symbolic parsers, to determine module structure.

Google’s Pathways architecture explores a different hybridization strategy by retaining a single large model while enforcing conditional sparse activation. In Pathways-enabled systems, only a subset of parameters is activated per input, approximating aspects of specialized processing. Switch Transformers and related Mixture-of-Experts (MoE) models instantiate this idea via learned routing over expert sub-networks, achieving strong performance with reduced per-example computation \citep{Fedus2021Switch}. In these models, modularity primarily emerges implicitly at the layer or expert level, while training and inference remain globally end-to-end. Unlike MoE systems, where specialization is largely motivated by computational efficiency, our architecture treats specialization as a representational and inferential prior, with explicit semantic interfaces and recurrent hypothesis validation.

A further class of hybrid approaches includes neuro-symbolic architectures, which integrate neural perception with explicit symbolic reasoning or structured knowledge representations \citep{Yi2018NSVQA, Mao2019NSCL}. By combining statistical learning with symbolic constraints, these systems offer improved interpretability and leverage prior knowledge, aligning with cognitive theories of structured reasoning \citep{Besold2017NeuralSymbolic}. However, they often rely on hand-designed symbolic components or task-specific supervision, limiting scalability.

Relatedly, predictive representation learning frameworks such as Joint Embedding Predictive Architectures (JEPA) emphasize learning abstract world models through prediction rather than reconstruction \citep{LeCun2022JEPA}. While JEPA-style models are typically monolithic, their emphasis on internal predictive structure resonates with our focus on hypothesis-driven, feedback-mediated inference. Our proposal extends this predictive perspective by embedding it within an explicitly modular, cortically inspired architecture.

\section{Diagnostic Proof-of-Concept Study}
\subsection{Overview and Architectural Alignment}
To provide an empirical anchor for the proposed cortically inspired modular architecture, we conduct a focused proof-of-concept (PoC) study examining whether explicit modular decomposition sharpens latent semantic structure within an existing monolithic model. While the full architecture described in Section~3 involves specialist encoders, routing controllers, and recurrent predictive feedback, the present experiment isolates a single representational component: \emph{modular factorization of latent features}. The goal is not to replicate the full architecture, but to test whether semantic partitioning alone alters the organization, concentration, and stability of internal representations.
This experiment should therefore be interpreted as a diagnostic proxy for the \textbf{modular specialization principle} of the broader architecture rather than a complete architectural instantiation. Please note complete methodology, statistical controls, and extended limitations appear in Appendix-A.1. Figures 2-4 in the appendix visualize training convergence, domain clustering, and modular improvements.

\textbf{Experimental Design.} We train sparse autoencoders (SAEs) to factorize Mistral-7B layer-15 activations (4096-dimensional residual stream, final-token hidden states) from 200 prompts spanning four semantic domains (vision,  language, cross-modal, reasoning; 50 prompts each). While this dataset is small relative to typical SAE training regimes, the goal of this experiment is diagnostic rather than benchmark-oriented. Two conditions are compared:

\textbf{Monolithic.} Single SAE (4096→1024→4096) trained on all prompts

\textbf{Modular.} Four domain-specific SAEs (4096→256→4096 each) with ground-truth semantic routing. We additionally train a capacity-matched monolithic control (256 features) to isolate modularity effects from representational budget constraints.

\textbf{Key Findings:} Modular decomposition produces three interpretability-relevant effects:

\begin{enumerate}
\item \textbf{Within-domain stability (+15.4pp).} Jaccard overlap of active features rises from 55.7\% (monolithic, within-domain) to 71.1\% (modular), indicating more consistent feature usage. This improvement is robust across all four domains: vision (+15.0pp), language (+3.8pp), cross-modal (+17.4pp), reasoning (+25.4pp).

\item \textbf{Semantic clustering (modest).} Observed feature-domain entropy (3.23) is significantly lower than the 100-run shuffled baseline (3.52 $\pm$ 0.01; p < 0.01), indicating reliable concentration. However, capacity-matched control (entropy=2.70) reveals this effect is partially 
explained by representational budget rather than pure modularity. Critically, reduced capacity alone does not account for the substantial within-domain stability gains (+15.4pp), which occur even at matched capacity.

\item \textbf{Feature specialization (minimal):} Only 6.2\% of features are domain-exclusive vs. 5.0\% $\pm$ 1.0\% random baseline, suggesting features remain largely distributed. Reconstruction fidelity is preserved (MSE: 0.0026 vs. 0.0031).
\end{enumerate}

\textbf{Interpretation.} Modular decomposition primarily enhances \emph{within-domain consistency} rather than enforcing \emph{hard feature partitioning}. This aligns with neuroscientific accounts of cortical specialization, where functional regions exhibit biased activation without strict exclusivity. The +15pp stability improvement is the primary robust effect and suggests that explicit semantic decomposition can bias representations toward greater within-domain consistency even with ground-truth routing—motivating investigation of learned routing mechanisms and end-to-end architectural benefits.

\textbf{Scope and Limitations.} This diagnostic tests only latent feature factorization, not the full proposed architecture (specialist encoders, learned routing, predictive feedback). Key constraints include: (1) capacity confound partially explains entropy effects, (2) small sample size (50/domain) limits generalization claims, (3) ground-truth semantic labels isolate decomposition effects from routing optimization. This choice intentionally upper-bounds the effect of idealized modular routing, establishing whether modularization is even worth learning, and (4) effect sizes are modest for entropy and specialization.

\section{Conclusion}
We have argued that perceptual AI can benefit substantially from three core cortical principles—modular specialization, predictive feedback, and shared latent spaces - which together support interpretable, adaptive, and robust systems. Decomposing perception into specialized expert units makes errors traceable and correctable at their source. Recurrent feedback loops allow what are traditionally labeled ``hallucinations" to function instead as provisional hypotheses, iteratively refined or rejected through internal validation. A shared cross-modal latent space further enables flexible, zero-shot reasoning across vision, language, and other modalities.

Such an architecture is particularly well suited to real-world and embodied settings. When sensors fail or inputs are incomplete - as in autonomous driving, robotics, or environmental monitoring - other modules can compensate, while top-down priors support principled inference over missing information, much like human perception reconstructs coherent scenes from partial evidence. By enabling redundancy, introspection, and controlled inference, modularity transforms fragility into resilience.

Realizing this vision will require new benchmarks that reward composability and feedback-driven refinement, tools for dynamic routing and module-level auditing, and deeper integration between AI research and neuroscience. Bridging these domains offers a path toward perceptual systems that are not only more capable, but also more transparent, robust, and cognitively grounded.

\bibliography{iclr2026_conference}
\bibliographystyle{iclr2026_conference}

\appendix
\section{Appendix}
\subsection{Proof-of-Concept Full Methodology}
\subsubsection{Dataset: Semantic Prompt Set}
We construct a balanced prompt dataset of 200 short text inputs spanning four semantic domains - \textbf{vision, language, cross-modal, and reasoning} - with 50 prompts per domain. Each domain contains a mixture of concrete descriptive, abstract explanatory, procedural, and edge or unusual variants in order to elicit heterogeneous internal activations rather than optimize for task performance. The dataset is designed to probe representational diversity and semantic clustering, not downstream accuracy.

\subsubsection{Model and Activation Extraction}
\textbf{Model.} We use the pretrained \texttt{Mistral-7B-v0.1} causal language model in evaluation mode with FP16 precision and no fine-tuning. All random seeds are fixed (seed=42) for reproducibility.

\textbf{Activation Extraction.} Internal representations are collected from the residual stream output of transformer block \texttt{layer 15} using a forward hook on \texttt{model.model.layers[layer\_idx]}. We extract the final-token hidden state for each prompt. This captures the integrated block output following self-attention, MLP, and residual updates rather than a single sublayer write, providing a stable representational snapshot. If no activations are collected, execution halts to avoid silent fallbacks.

\subsubsection{Sparse Autoencoder Factorization} We employ sparse autoencoders (SAEs) as lightweight interpretable factorization mechanisms over the 4096-dimensional activation space.

\textbf{Monolithic SAE.} A single encoder–decoder pair with ReLU nonlinearity (4096 $\rightarrow$ 1024 $\rightarrow$ 4096) is trained on the full activation set.

\textbf{Capacity-Matched Control.} To isolate modularity effects from representational budget, we additionally
train a reduced monolithic SAE (4096 $\rightarrow$ 256 $\rightarrow$ 4096).

\textbf{Per-Expert SAEs.} For the modular condition, activations are partitioned by semantic domain and four independent SAEs are trained with reduced capacity (4096 $\rightarrow$ 256 $\rightarrow$ 4096) on $\sim$50 samples each. Routing is based on ground-truth semantic labels rather than learned gating in order to isolate the effect of decomposition itself.

\textbf{Training Details.} Optimizer: Adam; learning rate $10^{-3}$; epochs: 20; batch size: 8; sparsitypenalty: L1 with $\lambda = 0.01$; gradient clipping norm 1.0; implementation: PyTorch on a single T4 GPU.

\subsubsection{Detailed Metric Definitions}
\begin{itemize}
\item \textbf{Sparsity Metrics:} Mean active features per sample and sparsity fraction, measuring representational efficiency.
\item \textbf{Domain–Feature Alignment:} Frequency of top-activated features aggregated by semantic domain to identify domain-specific activation patterns.
\item \textbf{Entropy Concentration:} Shannon entropy of feature distributions across domains; lower entropy indicates features concentrate in fewer dimensions per domain.

\item \textbf{Negative Control:} Shuffled-label baseline repeated 100 times with different random seeds to estimate null entropy distribution and compute $p$-values.

\item \textbf{Multi-Seed Stability:} Jaccard overlap of top features across five independent SAE trainings to verify features are not random artifacts.

\item \textbf{Within-Domain Stability:} Pairwise Jaccard overlap of active feature sets within each semantic domain; higher overlap indicates consistent feature usage across samples in the same domain.

\item \textbf{Feature Specialization (Shared-Space Exclusivity):} Fraction of top features (measured in a shared monolithic SAE feature space) that activate uniquely within a single domain; higher values indicate domain-exclusive
representations. Measured in shared space to enable fair comparison with random routing baseline.

\item \textbf{Reconstruction Trade-off:} Mean squared reconstruction error to ensure modular decomposition does not degrade representational fidelity.
\end{itemize}

\subsubsection{Results}
Modular decomposition substantially improves within-domain stability (+15.4pp) while maintaining comparable reconstruction fidelity as indicated in Table 1 and visualized in Figures 2-4. Figure 2 demonstrates stable training convergence, Figure 3 shows clear domain-specific clustering patterns, and Figure 4 illustrates the primary stability improvement (+15.4pp) alongside modest specialization gains. Capacity-matched control shows entropy reductions are partially explained by representational budget constraints.

\textbf{Note.} Entropy is not reported for the per-expert condition because features are partitioned across four independent 256-dimensional SAEs. Within-domain stability is not measured for the capacity-matched control. The entropy reduction (2.70 vs.\ 3.23) indicates that reduced representational budget alone increases feature concentration, partially confounding entropy-based interpretations. Random routing baseline yields $5.0\% \pm 1.0\%$ specialization versus $6.2\%$ under ground-truth semantic routing (measured in a shared monolithic feature space).

\begin{table}[t]
\caption{Comparison of monolithic and modular SAE factorization conditions with statistical controls. }

\label{tab:poc_detailed}
\begin{center}
\footnotesize
\setlength{\tabcolsep}{6pt}
\begin{tabular}{lccc}
\multicolumn{1}{c}{\bf Condition} 
& \multicolumn{1}{c}{\bf Stability} 
& \multicolumn{1}{c}{\bf Entropy} 
& \multicolumn{1}{c}{\bf MSE} \\
\hline \\

Monolithic (1024 features) 
& 0.557 
& 3.23 
& 0.0026 \\

Per-Expert (4$\times$256, ground-truth routing) 
& \textbf{0.711} 
& -- 
& 0.0031 \\

\\
\multicolumn{4}{l}{\emph{Controls and Baselines}} \\
\\

Shuffled Labels (100 runs) 
& -- 
& $3.52 \pm 0.01$ 
& -- \\

Capacity-Matched Monolithic (256) 
& -- 
& 2.70 
& 0.0033 \\

\end{tabular}

\vspace{0.5em}
\raggedright

\end{center}
\end{table}

\begin{figure}[h]
    \centering
    \includegraphics[width=\linewidth]
    {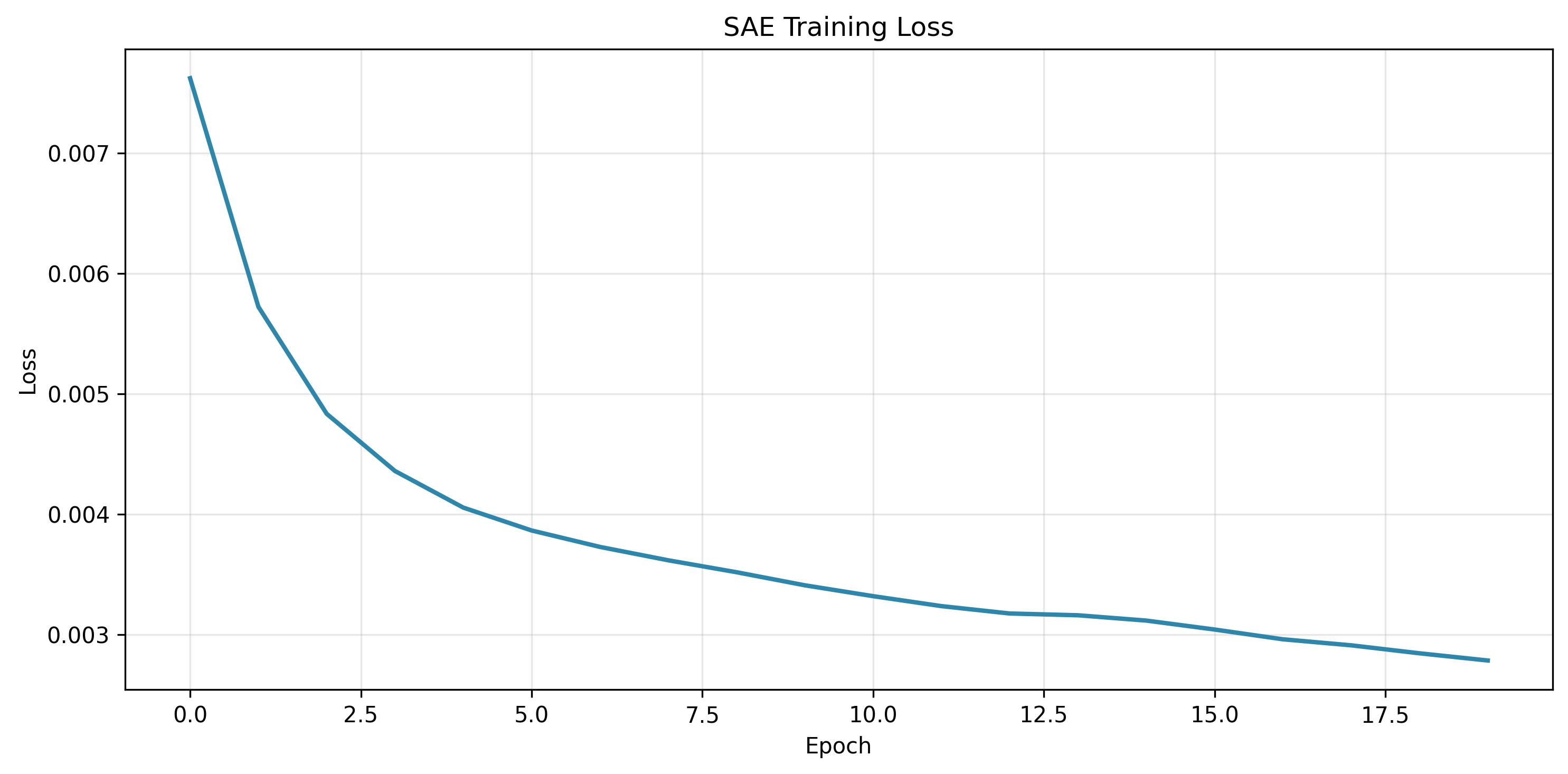}
    \caption{SAE training convergence (MSE=0.0026 at epoch 20).}
    \label{fig:results_1}
\end{figure}

\begin{figure}[h]
    \centering
    \includegraphics[width=\linewidth]
    {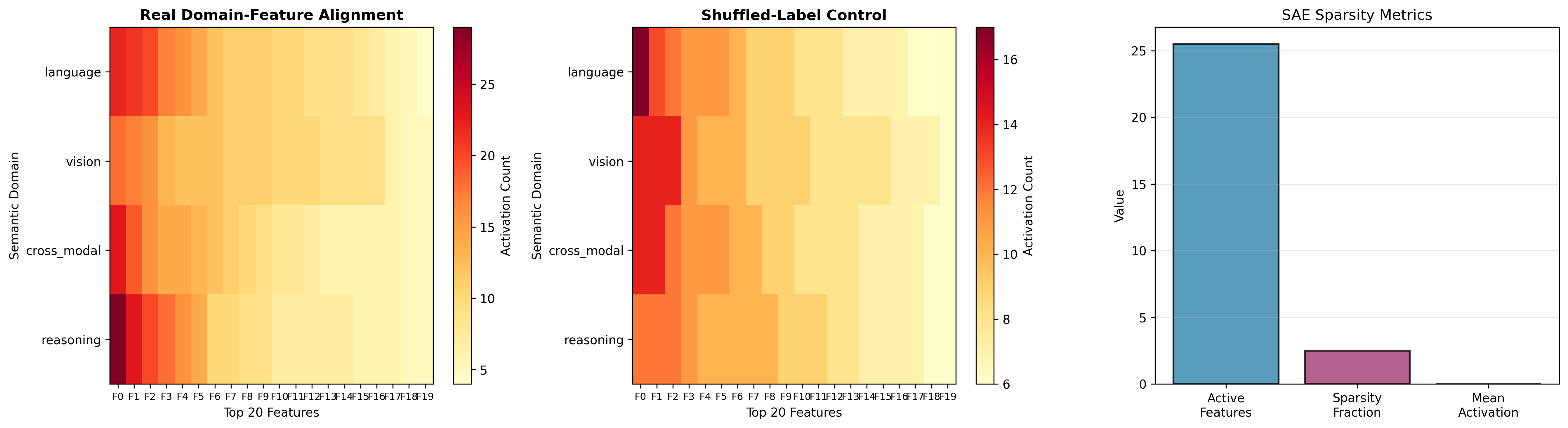}
    \caption{Domain-feature clustering: real (left), shuffled control (center), sparsity metrics (right)}.
    \label{fig:results_3}
\end{figure}

\begin{figure}[h]
    \centering
    \includegraphics[width=\linewidth]
    {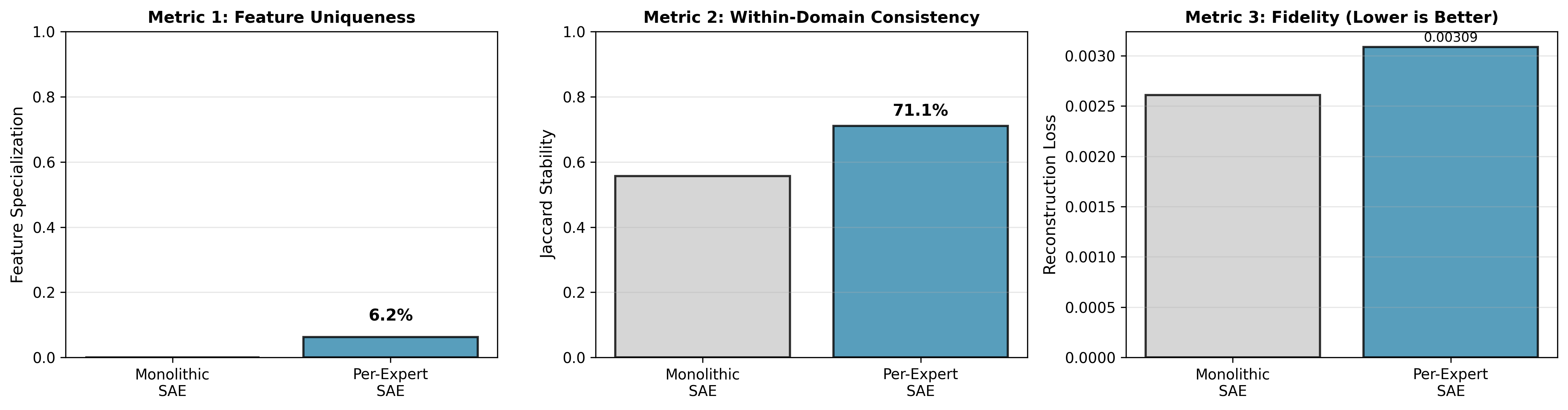}
    \caption{Modular factorization: feature uniqueness (left), within-domain stability +15.4pp (center), reconstruction fidelity (right).}
    \label{fig:results_2}
\end{figure}

\paragraph{Sparse and Domain-Concentrated Representations.} The monolithic SAE produces highly sparse encodings ($\sim$25 active features per sample; $\sim$2.5\% sparsity fraction). Feature activation patterns exhibit clear semantic clustering: entropy of the real domain–feature distribution (3.23) is significantly lower than the 100-run shuffled baseline (3.52 $\pm$ 0.01, $p < 0.01$; 100-run permutation test), indicating statistically reliable, though moderate, non-random concentration of features by domain. However, the capacity-matched monolithic SAE (256 features)
achieves entropy 2.70, lower than the standard condition, indicating that capacity reduction alone increases concentration independent of modularity. This partially confounds the entropy interpretation and suggests the effect reflects both semantic structure and representational budget constraints.

\paragraph{Stability Across Training Runs.} Multi-seed analysis yields a mean top-feature Jaccard overlap of $0.364 \pm 0.014$ across five independent trainings, suggesting moderate but consistent structural regularities rather than purely random feature allocations.

\paragraph{Modular Factorization Effects.}
When encoded through a shared monolithic SAE feature space (to enable fair comparison with random routing), per-expert specialization is modest: 6.2\% unique features versus 5.0\% $\pm$ 1.0\% random baseline (10 runs). This indicates that most features remain distributed rather than strictly partitioned by domain. However, modular decomposition substantially increases \emph{within-domain stability}: average Jaccard overlap rises from 0.557 (monolithic, evaluated within-domain) to 0.711 (per-expert), an absolute improvement of 15.4 percentage points (pp). This improvement is consistent across all four domains: vision (+15.0pp), language (+3.8pp), cross-modal (+17.4pp), and reasoning (+25.4pp). Reconstruction loss remains comparable between conditions (monolithic MSE=0.0026, per-expert=0.0031), indicating that increased structural consistency does not meaningfully degrade fidelity.

\paragraph{Interpretation.} Taken together, these results suggest that modular decomposition primarily sharpens within-domain reliability and secondarily increases concentration of shared features rather than enforcing hard representational separation. Features remain largely distributed, but become more stable and semantically coherent under explicit partitioning. The primary robust effect is improved within-domain consistency (+15.4pp Jaccard), while entropy and specialization effects are modest and partially confounded by capacity constraints. This pattern aligns with neuroscientific accounts of cortical specialization, where functional regions exhibit biased activation and recurrent stabilization without strict exclusivity.

\subsubsection{Detailed Limitations and Scope}

This PoC study has several constraints that position it as exploratory rather than confirmatory.

\paragraph{Capacity Confound.} We reduce per-expert SAE capacity (256 vs. 1024 monolithic) to prevent overparameterization in the 50-sample regime. However, the capacity-matched control reveals that reduced capacity alone increases entropy concentration (2.70 vs. 3.23), indicating the entropy effect is partially explained by representational budget constraints rather than pure modularity. While per-expert SAEs show superior within-domain stability at matched capacity, the entropy and specialization metrics are confounded. Future work should match
per-sample capacity (e.g., 4$\times$256-feature experts vs. 1024-feature monolithic per-sample capacity) to fully isolate modularity effects.

\paragraph{Sample Size and Generalization.} With only 50 samples per expert, overfitting risk exists despite capacity reduction and sparsity regularization. We evaluate SAEs on the same activations used for training—standard practice in interpretability research—but cannot validate whether learned features generalize to novel prompts within each domain. Cross-validation with larger prompt pools (200--500 per domain) is needed to confirm domain-general feature extraction.

\paragraph{Ground-Truth Semantic Routing.} We use ground-truth semantic labels rather than learned routing to isolate decomposition effects from routing optimization. Real deployments require learning routing functions, which may not achieve perfect semantic alignment and could introduce failure modes not captured in this controlled setting. Investigating learned routing mechanisms (e.g., attention-based gating, gradient-based assignment) is essential for end-to-end architectural validation.

\paragraph{Effect Size Magnitude.}
While statistically robust ($p < 0.01$, validated across 100 shuffled baselines), entropy differences are modest ($\Delta=0.30$, 9\% reduction). Feature specialization shows minimal domain-exclusivity (6.2\% vs. 5.0\% random baseline). The primary robust effect is within-domain stability improvement (+15.4pp), suggesting modularity enhances consistency more than exclusivity. This indicates features are \emph{biased} toward domains rather than \emph{exclusive} to them.

\paragraph{Measurement Framework.}
Comparison metrics measure different structural properties. Jaccard stability assesses local consistency within domains, while specialization measures global exclusivity across domains. These are complementary but not directly comparable. Additionally, feature interpretability to humans (e.g., via manual inspection or probe classifiers) remains unvalidated; automated metrics do not guarantee semantic meaningfulness or causal disentanglement.

\paragraph{Architectural Scope.}
This tests only latent feature factorization. Specialist encoders, routing controllers, and predictive feedback mechanisms remain unvalidated. The PoC demonstrates that explicit decomposition can improve representational
properties under ground-truth routing but does not validate end-to-end architectural benefits such as emergent compositionality, transfer learning, or catastrophic forgetting mitigation.

These constraints necessitate larger-scale replication with learned routing, held-out evaluation, diverse prompt distributions, and capacity-matched comparisons before drawing architectural conclusions. The present work establishes preliminary evidence that modular decomposition enhances within-domain feature consistency while highlighting methodological requirements for stronger validation.

\end{document}